\documentclass{article} 
\usepackage{iclr2020_conference,times}

\usepackage{hyperref}
\usepackage{url}
\usepackage{subfig}

\usepackage{algorithm}
\usepackage{algorithmic}

\usepackage{graphicx}
\graphicspath{{figures/}}
\usepackage{amsmath}
\usepackage{amssymb}

\DeclareMathOperator*{\argmax}{arg\,max}

\iclrfinalcopy

\title{Robust Visual Domain Randomization for Reinforcement Learning}

\author{Reda Bahi Slaoui$^1$, William R. Clements$^1$, Jakob N. Foerster$^2$, S\'{e}bastien Toth$^1$ \\
$^1$Indust.ai, Paris, France \\
$^2$Facebook Artificial Intelligence Research \\}

\newcommand\blfootnote[1]{%
  \begingroup
  \renewcommand\thefootnote{}\footnote{#1}%
  \addtocounter{footnote}{-1}%
  \endgroup
}

\newtheorem{definition}{Definition}
 
 \newtheorem{prop}{Proposition}
  \newtheorem{lemma}{Lemma}

\begin{document}

\maketitle

\begin{abstract}
Producing agents that can generalize to a wide range of visually different environments is a significant challenge in reinforcement learning. One method for overcoming this issue is visual domain randomization, whereby at the start of each training episode some visual aspects of the environment are randomized so that the agent is exposed to many possible variations. However, domain randomization is highly inefficient and may lead to policies with high variance across domains. Instead, we propose a regularization method whereby the agent is only trained on one variation of the environment, and its learned state representations are regularized during training to be invariant across domains. We conduct experiments that demonstrate that our technique leads to more efficient and robust learning than standard domain randomization, while achieving equal generalization scores.
\end{abstract}

Overfitting to a training environment is a serious issue in reinforcement learning. Agents trained on one domain often fail to generalize to other domains that differ only in small ways from the original domain \citep{sutton1996generalization,cobbe2018quantifying,DBLP:journals/corr/abs-1804-06893,DBLP:journals/corr/abs-1810-12282,DBLP:journals/corr/abs-1806-07937,DBLP:journals/corr/abs-1812-02868,DBLP:journals/corr/abs-1810-00123}. Good generalization is essential for fields such as robotics and autonomous vehicles, where the agent is often trained in a simulator and is then deployed in the real world where novel conditions will certainly be encountered. Transfer from such simulated training environments to the real world is known as crossing the \textit{reality gap} in robotics, and is well known to be difficult, thus providing an important motivation for studying generalization. 

We focus on the problem of generalizing between environments that visually differ from each other, for example in color or texture. Domain randomization, in which the visual and physical properties of the training domains are randomized at the start of each episode during training, has been shown to lead to improved generalization and transfer to the real world with little or no real world data \citep{Mordatch2015EnsembleCIOFD,DBLP:journals/corr/RajeswaranGLR16,DBLP:journals/corr/TobinFRSZA17, DBLP:journals/corr/SadeghiL16, DBLP:journals/corr/AntonovaCSK17, DBLP:journals/corr/abs-1710-06537, DBLP:journals/corr/abs-1808-00177}. However, domain randomization has been empirically shown to often lead to suboptimal policies with high variance in performance over different randomizations \citep{DBLP:journals/corr/abs-1904-04762}. This issue can cause the learned policy to underperform in any given target domain. 

Instead, we propose a regularization method whereby the agent is only trained on one variation of the environment, and its learned state representations are regularized during training to be invariant across domains. We provide theoretical justification for this regularization, and experimentally demonstrate that this method outperforms alternative domain randomization and regularization techniques for transferring the learned policy to domains within the randomization space.\blfootnote{Correspondence to: william.clements@indust.ai}

\section{Related Work}

We distinguish between two types of domain randomization: visual randomization, in which the variability between domains should not affect the agent's policy, and dynamics randomization, in which the agent should learn to adjust its behavior to achieve its goal. Visual domain randomization, which we focus on in this work, has been successfully used to directly transfer RL agents from simulation to the real world without requiring any real images \citep{DBLP:journals/corr/TobinFRSZA17, DBLP:journals/corr/SadeghiL16, DBLP:journals/corr/abs-1902-03701}. However, prior work has noted that domain randomization is inefficient \cite{DBLP:journals/corr/abs-1904-04762}. On the other hand, whereas the notion of learning similar state representations to improve transfer to new domains is not new \cite{DBLP:journals/corr/TzengDHFPLSD15,DBLP:journals/corr/GuptaDLAL17,DBLP:journals/corr/DaftryBH16}, previous work has focused on cases when simulated and target domains are both known, and cannot straightforwardly be applied when the target domain is only known to be within a distribution of domains.

We note that concurrently and independently of our work, \cite{aractingi2019improving} and \cite{lee2019simple} also propose regularization schemes for learning policies that are invariant to randomized visual changes in the environment without any real world data. In addition to proposing a similar scheme, we theoretically motivate and experimentally investigate the effects of our regularization on the learned features. Moreover, whereas \cite{aractingi2019improving} propose regularizing the network outputs, we regularize intermediate layers instead, which we show in the appendix to be beneficial. Also, \cite{lee2019simple} propose randomizing convolutional layer weights instead of directly perturbing input observations, as we do.

\section{Theory: Variance of a policy across randomized domains}

We consider Markov decision processes (MDP) defined by $(\mathcal{S}, \mathcal{A}, R, T, \gamma)$, where $\mathcal{S}$ is the state space, $\mathcal{A}$ the action space, $R : \mathcal{S} \times \mathcal{A} \rightarrow \mathbb{R}$ the reward function, $T : \mathcal{S} \times \mathcal{A} \rightarrow Pr(\mathcal{S})$ the transition dynamics, and $\gamma$ the discount factor. An agent is defined by a policy $\pi$ that maps states to distributions over actions, such that $\pi(a|s)$ is the probability of taking action $a$ in state $s$.

The visual domain randomization problem involves considering a set of functions $\{\phi\}$ that map one set of states to another set of states $\phi : \mathcal{S} \rightarrow \mathcal{S}'$, without affecting any other aspects of the MDP. Despite the fact that visual randomizations do not affect the underyling MDP, the agent's policy can still vary between domains. To compare the ways in which policies can differ between randomized domains, we introduce the notion of Lipschitz continuity of a policy over a set of randomizations.

\begin{definition}
We assume the state space is equipped with a distance metric. A policy $\pi$ is Lipschitz continuous over a set of randomizations $\{\phi\}$ if
$$ K_{\pi} = \sup_{\phi_1, \phi_2 \in \{\phi\}} \sup_{s\in \mathcal{S}} \frac{D_{TV}(\pi(\cdot|\phi_1(s)) \| \pi(\cdot|\phi_2(s)))}{ |\phi_1(s)-\phi_2(s)|}$$ 
is finite. Here, $D_{TV}(P\|Q)$ is the total variation distance between distributions.
\end{definition}

The following inequality shows that this Lipschitz constant is crucial in quantifying the robustness of RL agents over a randomization space. Informally, if a policy is Lipschitz continuous over randomized MDPs, then small changes in the background color in an environment will have a small impact on the policy. The smaller the Lipschitz constant, the less a policy is affected by different randomization parameters.

\begin{prop}
We consider a set of randomizations $\{\phi\}$ of the states of an MDP M. Let $\pi$ be a K-Lipschitz policy over $\{\phi\}$. Suppose the rewards are bounded by $r_{\max} $ such that $\forall a\in \mathcal{A}, s \in \mathcal{S}, |r(s,a)| \leq r_{\max}$.
Then for all $\phi_1$ and $\phi_2$ in $\{\phi\}$, the following inequalities hold :
\begin{align}
    |\eta_1 - \eta_2| \leq 2 r_{\max} \sum_t^\infty \gamma^t \min(1,(t+1)K_\pi\|\phi_1 - \phi_2\|_\infty) \leq \frac{2 r_{\max}K_\pi }{(1-\gamma)^2}\|\phi_1 - \phi_2\|_\infty
\end{align}
Where $\eta_i$ is the expected cumulative return of the policy on the MDP with states randomized by $\phi_i$, for $i \in \{1,2\}$, and $|\phi_1 - \phi_2\|_\infty = \sup_{s\in \mathcal{S}} |\phi_1(s) - \phi_2(s)|$.
\label{proposition}
\end{prop}

The proof for proposition 1 is presented in the appendix. Proposition 1 shows that reducing the Lipschitz constant reduces the maximum variations of the policy over the randomization space. 

\section{Proposed regularization}

Motivated by the theory described above, here we present a regularization technique for producing low-variance policies over the randomization space by minimizing the Lipschitz constant of definition 1. We start by choosing one variation of the environment on which to train an agent with a policy $\pi$ parameterized by $\theta$, and during training we minimize the loss
\begin{align}
\mathcal{L}(\theta) &= \mathcal{L}_{RL}(\theta) + \lambda \mathop{\mathbb{E}}_{s \sim \pi} \mathop{\mathbb{E}}_\phi \|f_\theta(s) - f_\theta(\phi(s))\|_2^2
\label{equation:reg_loss}
\end{align} 
\noindent where $\lambda$ is a regularization parameter, $\mathcal{L}_{RL}$ is the loss corresponding to the chosen reinforcement learning algorithm, the first expectation is taken over the distribution of states visited by the current policy, and $f_\theta$ is a feature extractor used by the agent's policy. In our experiments, we choose the output of the last hidden layer of the value or policy network as our feature extractor. The regularization term in the loss causes the agent to learn representations of states that ignore variations caused by the randomization. 

In appendix C, we provide evidence on a toy problem that this regularization does indeed control for the Lipschitz constant of the policy and helps in lowering the theoretical bound in proposition 1.

\section{Experiments}

In the following, we experimentally demonstrate that our regularization leads to improved generalization compared to standard domain randomization with both value-based and policy-based reinforcement learning algorithms. Implementation details can be found in the appendix, and the code is available at \url{https://github.com/IndustAI/visual-domain-randomization}.

\subsection{Visual Cartpole with DQN}

We first consider a version of the OpenAI Gym Cartpole environment \citep{gym} in which we randomize the color of the background. For training, we use the DQN algorithm with a CNN architecture similar to that used by \cite{mnih2015human}. We compare the performance of three agents. The \textbf{Normal} agent is trained on the reference domain with a white background. The \textbf{Randomized} agent is trained on a specified randomization space $\{\phi\}$. The \textbf{Regularized} agent is trained on a white background using our regularization method with respect to randomization space $\{\phi\}$. The training of all three agents is done using the same hyperparameters, and over the same number of steps.

\begin{figure}[h]
\centering
\includegraphics[scale=0.27]{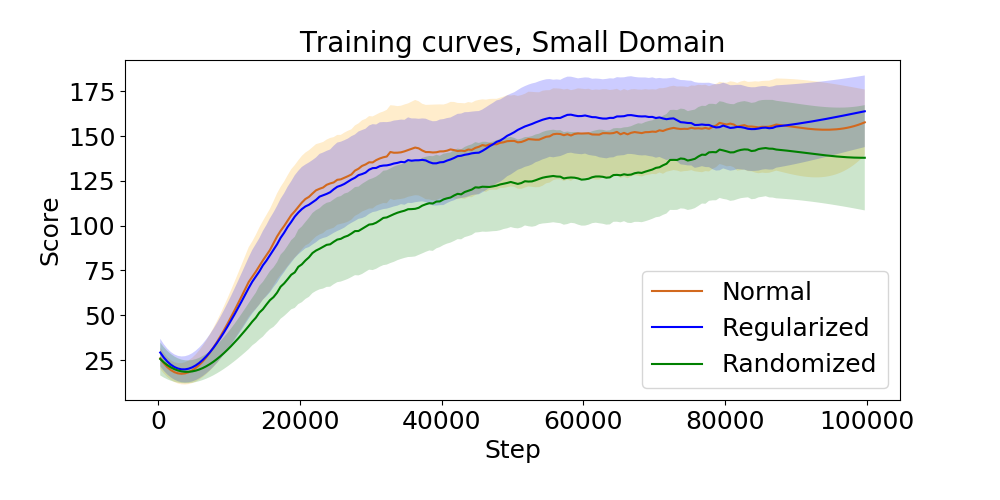}
\includegraphics[scale=0.27]{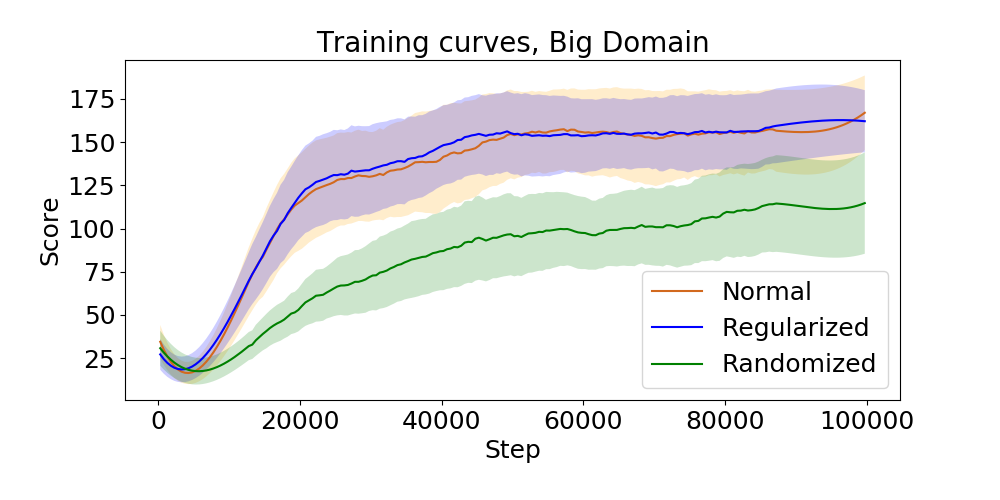}
\caption{Training curves over randomization spaces $\{\phi\}_{small}$ (left) and $\{\phi\}_{big}$ (right). Shaded areas indicate the 95\% confidence interval of the mean, obtained over 10 training seeds.}
\label{fig:trainingcurves}
\end{figure}

Figure \ref{fig:trainingcurves} shows the training curves of all three agents over two randomization spaces of different sizes: $\{\phi\}_{small}$ contains 1/8th of all colors in the RGB cube, and $\{\phi\}_{big}$ contains half the RGB cube. We find that the normal and regularized agents have similar training curves and the regularized agent is not affected by the size of the randomization space. However, the randomized agent learns more slowly on the small randomization space $\{\phi\}_{small}$ (left), and also achieves worse performance on the bigger randomization space $\{\phi\}_{big}$ (right). This indicates that standard domain randomization scales poorly with the size of the randomization space  $\{\phi\}$, whereas our regularization method is more robust to a larger randomization space.

\begin{figure}[h]
\centering
\includegraphics[scale=0.3]{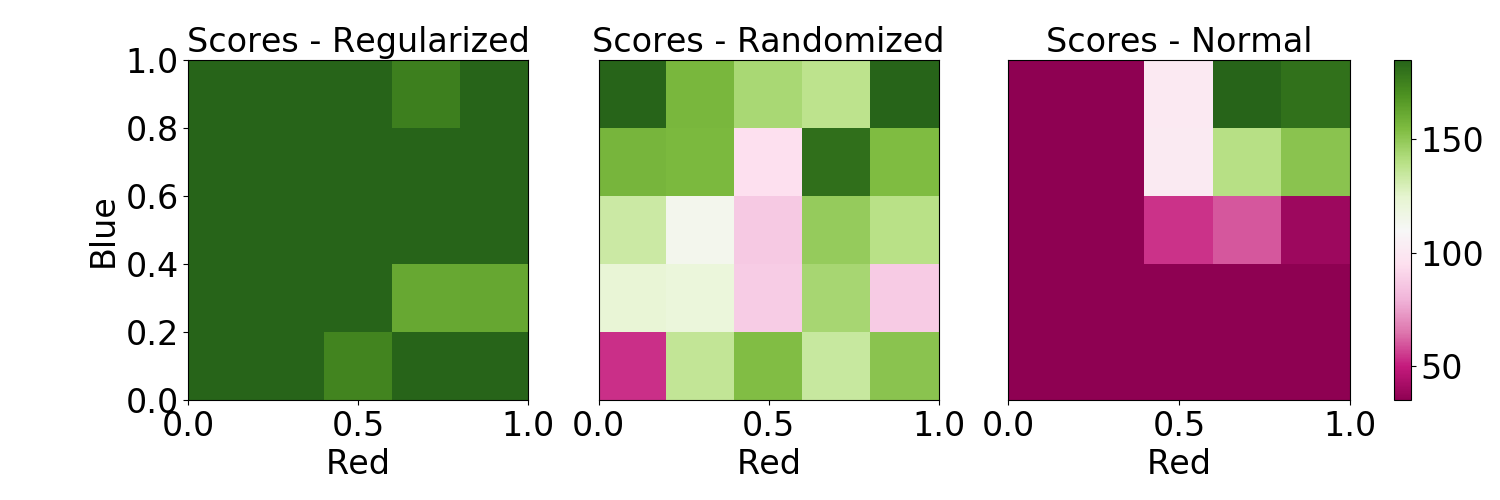}
\caption{Comparison of the average scores of different agents over different domains. The scores are calculated over a plane of the (r,g,b) where $g=1$ is fixed, averaged over 1000 steps. The training domain for both the regularized and normal agents is located at the top right. The regularized agent learns more stable policies than the randomized agent over these domains.}
\label{fig:interpolation}
\end{figure}

We now compare the returns of the policies learned by the agents in different domains within the randomization space. We select a plane within $\{\phi\}_{big}$ obtained by varying only the R and B channels but keeping G fixed. We plot the scores obtained on this plane in figure \ref{fig:interpolation}. We see that despite having only been trained on one domain, the regularized agent achieves consistently high scores on the other domains. On the other hand, the randomized agent's policy exhibits returns with high variance between domains, which indicates that different policies were learned for different domains. To explain these results, in the appendix we study the representations learned by the agents on different domains and show that the regularized agent learns similar representations for all domains while the randomized agent learns different representations.

\subsection{Car Racing with PPO}

\begin{figure}[ht]
\centering
\includegraphics[scale=0.15]{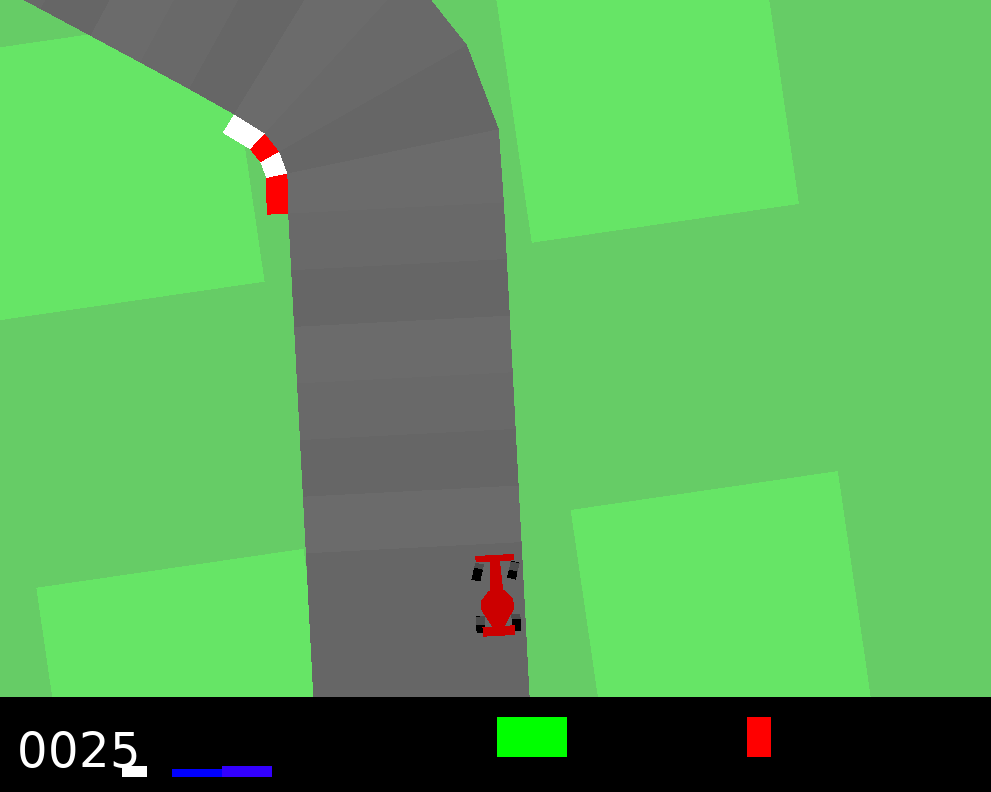}
\includegraphics[scale=0.15]{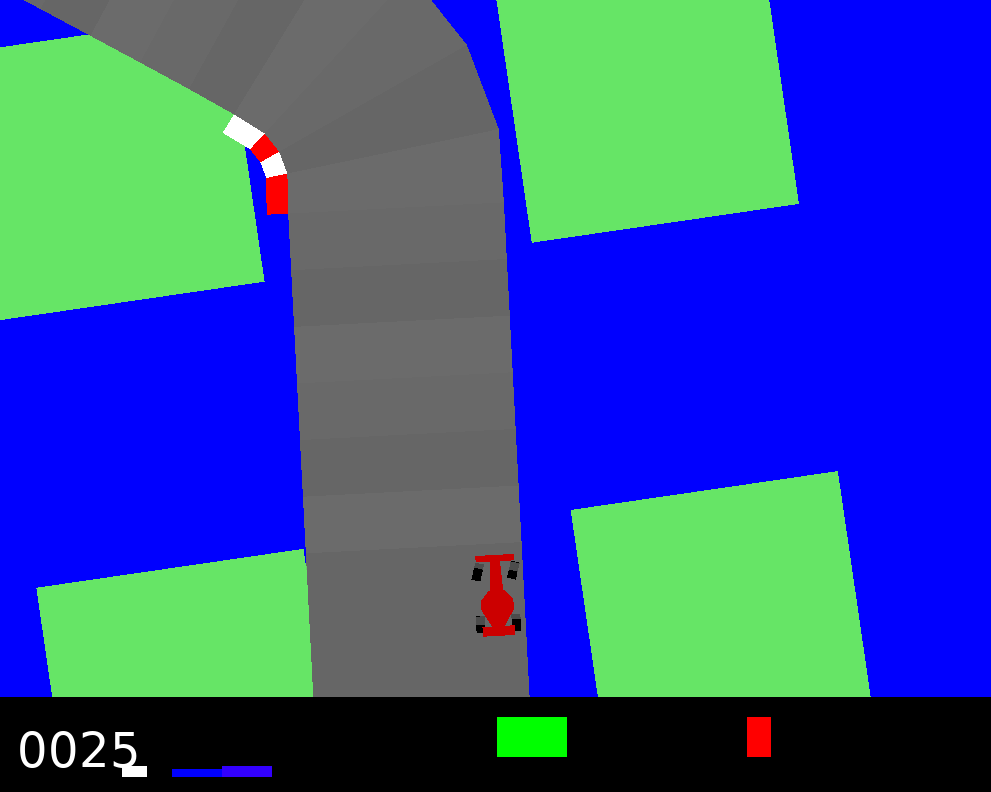}
\raisebox{0.7\height}{
\footnotesize{
\begin{tabular}{l|l|l}
    Agent & Return & Return \\ 
    & (original) & (all colors)  \\ \hline
    Normal & $554\pm68$ & $60\pm53$ \\
    Randomized & $-17\pm33$ & $-35\pm9$ \\ \hline
    Dropout 0.1 & $696\pm107$ & $154\pm86$ \\
    Weight decay $10^{-4}$ & $692\pm55$ & $61\pm15$ \\
    EPOpt-PPO & $-14\pm28$ & $-7\pm47$ \\ \hline    Regularized ($\lambda=10$) & $622\pm81$ & $324\pm51$ \\
    Regularized ($\lambda=50$) & $640\pm40$ & $553\pm80$
\end{tabular}
}}
\caption{Left: frames from the reference and a randomized CarRacing environment. Right: Average returns on the original environment and its randomizations over all colors, with 95\% confidence intervals calculated from 5 training seeds.}
\label{fig:carracing}
\end{figure}

To demonstrate the applicability of our regularization method to other domains and algorithms, we also perform experiments with the PPO algorithm \citep{schulman2017proximal} on the CarRacing environment \citep{gym}, in which an agent must drive a car around a racetrack. An example state from this environment and a randomized version in which part of the background changes color are shown in figure \ref{fig:carracing}. Randomization in this experiment occurs over the entire RGB cube, which is larger than for the cartpole experiments. We train several agents: a normal agent on the reference domain, a randomized agent, and two agents with our regularization and different values of $\lambda$. To compare our method to other regularization techniques, we also train agents with dropout and weight decay on the reference domain with the parameters of \cite{aractingi2019improving}, and an agent with the EPOpt-PPO algorithm \cite{DBLP:journals/corr/RajeswaranGLR16}, where in our implementation the agent only trains on the randomized domains on which its score is worse than average.

Scores on both the reference domain and its randomizations are shown in \ref{fig:carracing}. We see that the randomized agent and EPOpt-PPO fail to learn on this large randomization space. Of the other regularization schemes that we tested, we find that although they do improve learning on the reference domain, only dropout leads to a small improvement in generalization. On the other hand, our regularization produces agents that are both successful in training and successfully generalize to a wide range of backgrounds. Moreover, a larger value of $\lambda$ yields higher generalization scores, which is consistent with our empirical study of our regularization scheme in appendix C. 

\section{Conclusion}

In this paper we studied generalization to visually diverse environments in deep reinforcement learning. We illustrated the inefficiencies of standard domain randomization, and proposed a theoretically grounded regularization method that leads to robust, low-variance policies that generalize well. We conducted experiments in different environments using both on-policy and off-policy algorithms that demonstrate the wide applicability of our method.

\bibliography{iclr2020_conference}

\begin{thebibliography}{26}
\providecommand{\natexlab}[1]{#1}
\providecommand{\url}[1]{\texttt{#1}}
\expandafter\ifx\csname urlstyle\endcsname\relax
  \providecommand{\doi}[1]{doi: #1}\else
  \providecommand{\doi}{doi: \begingroup \urlstyle{rm}\Url}\fi

\bibitem[Antonova et~al.(2017)Antonova, Cruciani, Smith, and
  Kragic]{DBLP:journals/corr/AntonovaCSK17}
Rika Antonova, Silvia Cruciani, Christian Smith, and Danica Kragic.
\newblock Reinforcement learning for pivoting task.
\newblock \emph{arXiv preprint arXiv:1703.00472}, 2017.

\bibitem[Aractingi et~al.(2019)Aractingi, Dance, Perez, and
  Silander]{aractingi2019improving}
Michel Aractingi, Christopher Dance, Julien Perez, and Tomi Silander.
\newblock Improving the generalization of visual navigation policies using
  invariance regularization.
\newblock \emph{ICML 2019 Workshop RL4RealLife}, 2019.

\bibitem[Brockman et~al.(2016)Brockman, Cheung, Pettersson, Schneider,
  Schulman, Tang, and Zaremba]{gym}
Greg Brockman, Vicki Cheung, Ludwig Pettersson, Jonas Schneider, John Schulman,
  Jie Tang, and Wojciech Zaremba.
\newblock Openai gym, 2016.

\bibitem[Cobbe et~al.(2019)Cobbe, Klimov, Hesse, Kim, and
  Schulman]{cobbe2018quantifying}
Karl Cobbe, Oleg Klimov, Chris Hesse, Taehoon Kim, and John Schulman.
\newblock Quantifying generalization in reinforcement learning.
\newblock \emph{International Conference on Machine Learning}, 2019.

\bibitem[Daftry et~al.(2016)Daftry, Bagnell, and
  Hebert]{DBLP:journals/corr/DaftryBH16}
Shreyansh Daftry, J~Andrew Bagnell, and Martial Hebert.
\newblock Learning transferable policies for monocular reactive mav control.
\newblock \emph{International Symposium on Experimental Robotics}, pp.\  3--11,
  2016.

\bibitem[Farebrother et~al.(2018)Farebrother, Machado, and
  Bowling]{DBLP:journals/corr/abs-1810-00123}
Jesse Farebrother, Marlos~C Machado, and Michael Bowling.
\newblock Generalization and regularization in dqn.
\newblock \emph{arXiv preprint arXiv:1810.00123}, 2018.

\bibitem[Gupta et~al.(2017)Gupta, Devin, Liu, Abbeel, and
  Levine]{DBLP:journals/corr/GuptaDLAL17}
Abhishek Gupta, Coline Devin, Yuxuan Liu, Pieter Abbeel, and Sergey Levine.
\newblock Learning invariant feature spaces to transfer skills with
  reinforcement learning.
\newblock \emph{International Conference on Learning Representations}, 2017.

\bibitem[Kang et~al.(2019)Kang, Belkhale, Kahn, Abbeel, and
  Levine]{DBLP:journals/corr/abs-1902-03701}
Katie Kang, Suneel Belkhale, Gregory Kahn, Pieter Abbeel, and Sergey Levine.
\newblock Generalization through simulation: Integrating simulated and real
  data into deep reinforcement learning for vision-based autonomous flight.
\newblock \emph{International Conference on Robotics and Automation}, 2019.

\bibitem[Kempka et~al.(2016)Kempka, Wydmuch, Runc, Toczek, and
  Ja{\'s}kowski]{kempka2016vizdoom}
Micha{\l} Kempka, Marek Wydmuch, Grzegorz Runc, Jakub Toczek, and Wojciech
  Ja{\'s}kowski.
\newblock Vizdoom: A doom-based ai research platform for visual reinforcement
  learning.
\newblock In \emph{2016 IEEE Conference on Computational Intelligence and Games
  (CIG)}. IEEE, 2016.

\bibitem[Lee et~al.(2019)Lee, Lee, Shin, and Lee]{lee2019simple}
Kimin Lee, Kibok Lee, Jinwoo Shin, and Honglak Lee.
\newblock A simple randomization technique for generalization in deep
  reinforcement learning.
\newblock \emph{arXiv preprint arXiv:1910.05396}, 2019.

\bibitem[Mehta et~al.(2019)Mehta, Diaz, Golemo, Pal, and
  Paull]{DBLP:journals/corr/abs-1904-04762}
Bhairav Mehta, Manfred Diaz, Florian Golemo, Christopher~J. Pal, and Liam
  Paull.
\newblock Active domain randomization.
\newblock \emph{Conference on Robot Learning}, 2019.

\bibitem[Mnih et~al.(2015)Mnih, Kavukcuoglu, Silver, Rusu, Veness, Bellemare,
  Graves, Riedmiller, Fidjeland, Ostrovski, et~al.]{mnih2015human}
Volodymyr Mnih, Koray Kavukcuoglu, David Silver, Andrei~A Rusu, Joel Veness,
  Marc~G Bellemare, Alex Graves, Martin Riedmiller, Andreas~K Fidjeland, Georg
  Ostrovski, et~al.
\newblock Human-level control through deep reinforcement learning.
\newblock \emph{Nature}, 518\penalty0 (7540):\penalty0 529, 2015.

\bibitem[Mordatch et~al.(2015)Mordatch, Lowrey, and
  Todorov]{Mordatch2015EnsembleCIOFD}
Igor Mordatch, Kendall Lowrey, and Emanuel Todorov.
\newblock Ensemble-{CIO}: Full-body dynamic motion planning that transfers to
  physical humanoids.
\newblock \emph{2015 IEEE/RSJ International Conference on Intelligent Robots
  and Systems (IROS)}, pp.\  5307--5314, 2015.

\bibitem[OpenAI et~al.(2020)]{DBLP:journals/corr/abs-1808-00177}
OpenAI et~al.
\newblock Learning dexterous in-hand manipulation.
\newblock \emph{The International Journal of Robotics Research}, 39\penalty0
  (1):\penalty0 3--20, 2020.

\bibitem[Packer et~al.(2018)Packer, Gao, Kos, Kr{\"a}henb{\"u}hl, Koltun, and
  Song]{DBLP:journals/corr/abs-1810-12282}
Charles Packer, Katelyn Gao, Jernej Kos, Philipp Kr{\"a}henb{\"u}hl, Vladlen
  Koltun, and Dawn Song.
\newblock Assessing generalization in deep reinforcement learning.
\newblock \emph{arXiv preprint arXiv:1810.12282}, 2018.

\bibitem[Peng et~al.(2018)Peng, Andrychowicz, Zaremba, and
  Abbeel]{DBLP:journals/corr/abs-1710-06537}
Xue~Bin Peng, Marcin Andrychowicz, Wojciech Zaremba, and Pieter Abbeel.
\newblock Sim-to-real transfer of robotic control with dynamics randomization.
\newblock \emph{2018 IEEE international conference on robotics and automation
  (ICRA)}, pp.\  1--8, 2018.

\bibitem[Rajeswaran et~al.(2017)Rajeswaran, Ghotra, Levine, and
  Ravindran]{DBLP:journals/corr/RajeswaranGLR16}
Aravind Rajeswaran, Sarvjeet Ghotra, Sergey Levine, and Balaraman Ravindran.
\newblock Epopt: Learning robust neural network policies using model ensembles.
\newblock \emph{International Conference on Learning Representations}, 2017.

\bibitem[Sadeghi \& Levine(2017)Sadeghi and
  Levine]{DBLP:journals/corr/SadeghiL16}
Fereshteh Sadeghi and Sergey Levine.
\newblock {(CAD)${^{2}}$RL}: Real single-image flight without a single real
  image.
\newblock \emph{Robotics: Science and Systems Conference}, 2017.

\bibitem[Schulman et~al.(2017)Schulman, Wolski, Dhariwal, Radford, and
  Klimov]{schulman2017proximal}
John Schulman, Filip Wolski, Prafulla Dhariwal, Alec Radford, and Oleg Klimov.
\newblock Proximal policy optimization algorithms.
\newblock \emph{arXiv preprint arXiv:1707.06347}, 2017.

\bibitem[Sutton(1996)]{sutton1996generalization}
Richard~S Sutton.
\newblock Generalization in reinforcement learning: Successful examples using
  sparse coarse coding.
\newblock In \emph{Advances in neural information processing systems}, pp.\
  1038--1044, 1996.

\bibitem[Sutton et~al.(2000)Sutton, McAllester, Singh, and
  Mansour]{sutton2000policy}
Richard~S Sutton, David~A McAllester, Satinder~P Singh, and Yishay Mansour.
\newblock Policy gradient methods for reinforcement learning with function
  approximation.
\newblock In \emph{Advances in neural information processing systems}, pp.\
  1057--1063, 2000.

\bibitem[Tobin et~al.(2017)Tobin, Fong, Ray, Schneider, Zaremba, and
  Abbeel]{DBLP:journals/corr/TobinFRSZA17}
Joshua Tobin, Rachel Fong, Alex Ray, Jonas Schneider, Wojciech Zaremba, and
  Pieter Abbeel.
\newblock Domain randomization for transferring deep neural networks from
  simulation to the real world.
\newblock \emph{IEEE/RSJ international conference on intelligent robots and
  systems (IROS)}, 2017.

\bibitem[Tzeng et~al.(2015)Tzeng, Devin, Hoffman, Finn, Peng, Levine, Saenko,
  and Darrell]{DBLP:journals/corr/TzengDHFPLSD15}
Eric Tzeng, Coline Devin, Judy Hoffman, Chelsea Finn, Xingchao Peng, Sergey
  Levine, Kate Saenko, and Trevor Darrell.
\newblock Towards adapting deep visuomotor representations from simulated to
  real environments.
\newblock \emph{arXiv preprint arXiv:1511.07111}, 2015.

\bibitem[Witty et~al.(2018)Witty, Lee, Tosch, Atrey, Littman, and
  Jensen]{DBLP:journals/corr/abs-1812-02868}
Sam Witty, Jun~Ki Lee, Emma Tosch, Akanksha Atrey, Michael Littman, and David
  Jensen.
\newblock Measuring and characterizing generalization in deep reinforcement
  learning.
\newblock \emph{arXiv preprint arXiv:1812.02868}, 2018.

\bibitem[Zhang et~al.(2018{\natexlab{a}})Zhang, Ballas, and
  Pineau]{DBLP:journals/corr/abs-1806-07937}
Amy Zhang, Nicolas Ballas, and Joelle Pineau.
\newblock A dissection of overfitting and generalization in continuous
  reinforcement learning.
\newblock \emph{arXiv preprint arXiv:1806.07937}, 2018{\natexlab{a}}.

\bibitem[Zhang et~al.(2018{\natexlab{b}})Zhang, Vinyals, Munos, and
  Bengio]{DBLP:journals/corr/abs-1804-06893}
Chiyuan Zhang, Oriol Vinyals, Remi Munos, and Samy Bengio.
\newblock A study on overfitting in deep reinforcement learning.
\newblock \emph{arXiv preprint arXiv:1804.06893}, 2018{\natexlab{b}}.

\end{thebibliography}
\bibliographystyle{iclr2020_conference}

\newpage
\appendix

\section{Proof of Proposition 1}

The proof presented in the following applies to MDPs with a discrete action space. However, it can straightforwardly be generalized to continuous action spaces by replacing sums over actions with integrals over actions. 

\noindent The proof uses the following lemma : 
\begin{lemma}
For two distributions $p(x, y) = p(x)p(y|x)$ and $q(x, y) = q(x)q(y|x)$, we can bound the total variation distance of the joint distribution : 
\begin{align*}
 D_{TV}(p(\cdot,\cdot) \| q(\cdot,\cdot)) &\leq D_{TV}(p(\cdot) \| q(\cdot)) \\ &+ \max_x D_{TV}(p(\cdot|x) \| q(\cdot|x)) 
\end{align*}
\end{lemma}
\textit{Proof of the Lemma}.

We have that :
\begin{align*}
 D_{TV}(p(\cdot,\cdot) \| q(\cdot,\cdot)) & = \frac{1}{2} \sum_{x,y}|p(x,y) - q(x,y)| \\
&= \frac{1}{2} \sum_{x,y}|p(x)p(y|x) - q(x)q(y|x)| \\ &=  \frac{1}{2} \sum_{x,y}|p(x)p(y|x) - p(x)q(y|x) \\ &+ (p(x) - q(x))q(y|x)|\\
&\leq \frac{1}{2} \sum_{x,y} p(x)|p(y|x)-q(y|x)|\\ & +|p(x) - q(x)|q(y|x) \\
&\leq \max_x D_{TV}(p(\cdot|x) \| q(\cdot|x)) \\ &+ D_{TV}(p(\cdot) \| q(\cdot))
\end{align*}
\textit{Proof of the proposition. } 

Let $p_{\phi_i}^t(s,a)$ be the probability of being in state $\phi_i(s)$ at time $t$, and executing action $a$, for $i=1,2$. Since both MDPs have the same reward function, we have by definition that $\eta_i = \sum_{s,a} \sum_t \gamma^t p_{\phi_i}^t(s,a)r_t(s,a)$, so we can write : 
\begin{align}
|\eta_1 - \eta_2| &\leq \sum_{s,a} \sum_t \gamma^t |p_{\phi_1}^t(s,a)- p_{\phi_2}^t(s,a)|r_t(s,a) \nonumber \\
                    & \leq r_{\max} \sum_{s,a} \sum_t \gamma^t |p_{\phi_1}^t(s,a)- p_{\phi_2}^t(s,a)| \nonumber \\
                    &= 2 r_{\max} \sum_t \gamma^t D_{TV}(p_{\phi_1}^t(\cdot,\cdot) \| p_{\phi_2}^t(\cdot,\cdot))
\label{eq:dtv}
\end{align}
But $ p_{\phi_1}^t(s,a) = p_{\phi_1}^t(s) \pi^{\phi_1}(a|s)$ and $ p_{\phi_2}^t(s,a) = p_{\phi_2}^t(s) \pi^{\phi_2}(a|s)$, Thus (Lemma 1) : 
\begin{align} D_{TV}(p_{\phi_1}^t(\cdot,\cdot) \| p_{\phi_2}^t(\cdot,\cdot)) &\leq D_{TV}(p_{\phi_1}^t(\cdot) \| p_{\phi_2}^t(\cdot)) \nonumber \\&+ \max_s D_{TV}(\pi^{\phi_1}(\cdot|s) \| \pi^{\phi_2}(\cdot|s)) \nonumber \\
 &\leq D_{TV}(p_{\phi_1}^t(\cdot) \| p_{\phi_2}^t(\cdot)) \nonumber \\&+ K_\pi \| \phi_1 -\phi_2 \|_\infty
 \label{eq:dtvk}
\end{align}
We still have to bound $D_{TV}(p_{\phi_1}^t(\cdot) \| p_{\phi_2}^t(\cdot))$. For $s \in S$ we have that : 
\begin{align*}
|p_{\phi_1}^t(s)- p_{\phi_2}^t(s)| &\leq \sum_{s'}|p_{\phi_1}(s_t=s|s')p_{\phi_1}^{t-1}(s') \\&- p_{\phi_2}(s_t=s|s')p_{\phi_2}^{t-1}(s')| \\
&= \sum_{s'}|p_{\phi_1}(s_t=s|s')p_{\phi_1}^{t-1}(s')\\& - p_{\phi_2}(s_t=s|s')p_{\phi_1}^{t-1}(s') \\&+ p_{\phi_2}(s_t=s|s')p_{\phi_1}^{t-1}(s')\\&- p_{\phi_2}(s_t=s|s')p_{\phi_2}^{t-1}(s')| \\
&\leq \sum_{s'}p_{\phi_1}^{t-1}(s')|p_{\phi_1}(s|s')-p_{\phi_2}(s|s')|\\& + p_{\phi_2}(s|s')|p_{\phi_1}^{t-1}(s')-p_{\phi_2}^{t-1}(s')|
\end{align*}

Summing over $s$ we have that \begin{align*}
D_{TV}(p_{\phi_1}^t(\cdot)||p_{\phi_2}^t(\cdot)) &\leq \frac{1}{2} \sum_s \mathbb{E}_{s' \sim p_{\phi_1}^{t-1}}[|p_{\phi_1}(s|s')\\&- p_{\phi_2}(s|s')|] \\  
&+ D_{TV}(p_{\phi_1}^{t-1}(\cdot)||p_{\phi_2}^{t-1}(\cdot))
\end{align*}

But by marginalizing over actions :  $ p_{\phi_1}(s|s') = \sum_a \pi^{\phi_1}(a|s')p_{\phi_1}(s|a,s')$, and using the fact that $p_{\phi_1}(s|a,s') = T_{\phi_1}(s|a,s') = T_{\phi_2}(s|a,s') = p_{\phi_2}(s|a,s'):= p(s|a,s')$, we have that \begin{align*}
    |p_{\phi_1}(s|s') - p_{\phi_2}(s|s')| &= |\sum_a p(s|a,s')(\pi^{\phi_1}(a|s') \\& - \pi^{\phi_2}(a|s'))| \\
    &\leq \sum_a p(s|a,s')|\pi^{\phi_1}(a|s')\\& - \pi^{\phi_2}(a|s')|
\end{align*}  
And using $\sum_s p(s|a,s') = 1$ we have that : 
\begin{align*}
    &\frac{1}{2} \sum_s\mathbb{E}_{s' \sim p_{\phi_1}^{t-1}}[|p_{\phi_1}(s|s') - p_{\phi_2}(s|s')| ]
    \\&\leq \frac{1}{2}\mathbb{E}_{s' \sim p_{\phi_1}^{t-1}}\sum_a [\sum_s p(s|a,s')]  |\pi^{\phi_1}(a|s')- \pi^{\phi_2}(a|s')| \\
    &\leq \max_{s'}D_{TV}(\pi^{\phi_1}(\cdot|s) \| \pi^{\phi_2}(\cdot|s)) \\
    &\leq  K_\pi \| \phi_1 -\phi_2 \|_\infty
\end{align*}
Thus, by induction, and assuming  $D_{TV}(p_{\phi_1}^0(\cdot)||p_{\phi_2}^0(\cdot)) = 0$ : $$  D_{TV}(p_{\phi_1}^t(\cdot)||p_{\phi_2}^t(\cdot)) \leq t  K_\pi \| \phi_1 -\phi_2 \|_\infty$$
Plugging this into inequality \ref{eq:dtvk}, we get
\begin{align*}
    D_{TV}(p_{\phi_1}^t(\cdot,\cdot) \| p_{\phi_2}^t(\cdot,\cdot)) \leq (t+1)  K_\pi \| \phi_1 -\phi_2 \|_\infty
\end{align*}
We also note that the total variation distance takes values between 0 and 1, so we have
$$D_{TV}(p_{\phi_1}^t(\cdot,\cdot) \| p_{\phi_2}^t(\cdot,\cdot)) \leq \min(1, (t+1)  K_\pi \| \phi_1 -\phi_2 \|_\infty)$$
Plugging this into inequality \ref{eq:dtv} leads to our first bound,
\begin{align*}
    |\eta_1 - \eta_2| &\leq 2r_{\max} \sum_t \gamma^t \min(1,(t+1) K_\pi \| \phi_1 -\phi_2 \|_\infty)
\end{align*}
Our second, looser bound can now be achieved as follows,
\begin{align*}
    |\eta_1 - \eta_2| &\leq 2r_{\max} \sum_t \gamma^t(t+1) K_\pi \| \phi_1 -\phi_2 \|_\infty \\
     |\eta_1 - \eta_2|&\leq \frac{2r_{\max}}{(1-\gamma)^2} K_\pi \| \phi_1 -\phi_2 \|_\infty
\end{align*}

\section{Comparison of our work with \cite{aractingi2019improving}}

\begin{figure}[ht]
\centering
\includegraphics[scale=0.5]{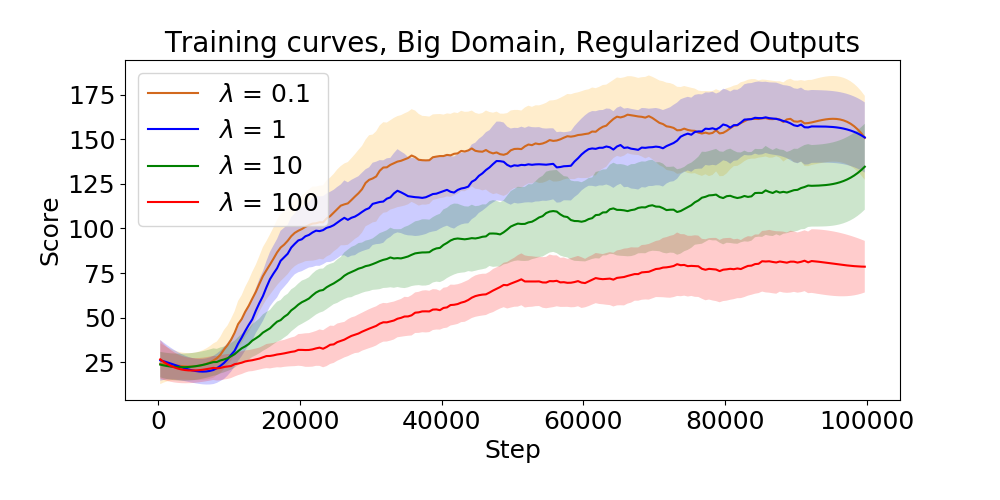}
\includegraphics[scale=0.3]{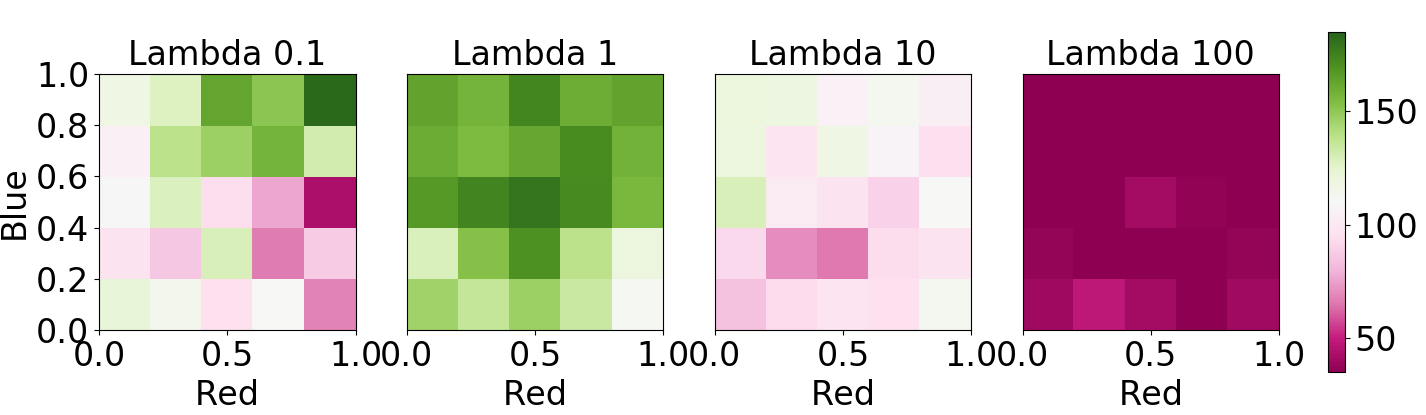}
\caption{Top: training curves of agents with different regularization strengths when following the scheme of \cite{aractingi2019improving}. Shaded errors correspond to 95\% confidence intervals of the mean, calculated from 10 training seeds. Bottom: scores obtained by trained agents for different regularization strengths on a plane within the RGB cube.}
\label{fig:outputreg}
\end{figure}

Concurrently and independently of our work, \cite{aractingi2019improving} propose a similar regularization scheme on randomized visual domains, which they experimentally demonstrate with the PPO algorithm on the VizDoom environment of \cite{kempka2016vizdoom} with randomized textures. As opposed to the regularization scheme proposed in our work in which we regularize the final hidden layer of the network, they propose regularizing the output of the policy network. Regularizing the last hidden layer as in our scheme more clearly separates representation learning and policy learning, since the final layer of the network is only trained by the RL loss. 

We hypothesized that regularizing the output of the network directly could lead to the regularization loss and the RL loss competing against each other, such that a tradeoff between policy performance and generalization would be necessary. To test this hypothesis, we performed experiments on the visual cartpole domain with output regularization with different values of regularization parameter $\lambda$. Our results are shown in figure \ref{fig:outputreg}. We find that increasing the regularization strength adversely affects training. However, agents trained with higher values of $\lambda$ do achieve more consistent results over the randomization space. This shows that there is indeed a tradeoff between generalization and policy performance when regularizing the network output as in \cite{aractingi2019improving}. In our experiments, however, we have found that changing the value of $\lambda$ only affects generalization ability and not agent performance on the reference domain.

\section{Impact of our regularization on the Lipschitz constant}

We conduct experiments on a simple gridworld to show that our regularization does indeed in general reduce the Lipschitz constant of the policy over randomized domains, leading to lower variance in the returns.

\begin{figure}[h]
\centering
\raisebox{10pt}{
\includegraphics[height=0.13\textheight]{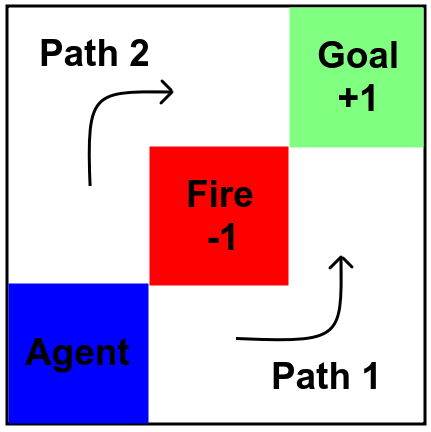}}
\qquad
\includegraphics[height=0.15\textheight]{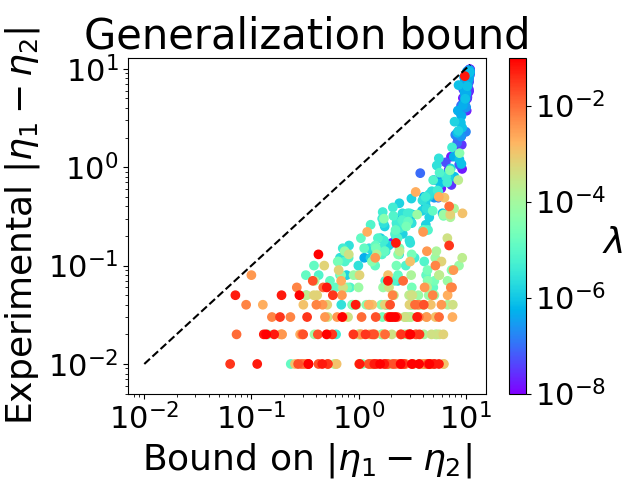}
\raisebox{50pt}{
\begin{tabular}{|c|c|}
  \hline
  Agent & Same path \\
  & probability \\
  \hline
Randomized& 86\% \\
  Regularized & \textbf{100\%} \\
   (ours) & \\
  \hline
\end{tabular}}
\caption{Left: a simple gridworld, in which the agent must make its way to the goal while avoiding the fire. Center: empirical differences between regularized agents' policies on two randomizations of the gridworld compared to our theoretical bound in equation \ref{proposition} (the dashed line). Each point corresponds to one agent, and 20 training seeds per value of $\lambda$ are shown here. Right: probability that different agents choose the same path for two randomizations of this domain. Our regularization method leads to more consistent behavior.}
\label{fig:gridworld}
\end{figure}

The environment we use is the $3\times3$ gridworld shown in figure \ref{fig:gridworld}, in which two optimal policies exist. The agent starts in the bottom left of the grid and must reach the goal while avoiding the fire. The agent can move either up or right, and in addition to the rewards shown in figure \ref{fig:gridworld} receives -1 reward for invalid actions that would case it to leave the grid. We set a time limit of 10 steps and $\gamma=1$. We introduce randomization into this environment by describing the state observed by the agent as a tuple $(x,y,\xi)$, where $(x,y)$ is the agent's position and $\xi$ is a randomization parameter with no impact on the underlying MDP. For this toy problem, we consider only two possible values for $\xi$: $+5$ and $-5$. The agents we consider use the REINFORCE algorithm \citep{sutton2000policy} with a baseline (see Algorithm 2), and a multi-layer perceptron as the policy network. 

First, we observe that even in a simple environment such as this one, a randomized agent regularly learns different paths for different randomizations (figure \ref{fig:gridworld}). An agent trained only on $\xi=5$ and regularized with our technique, however, consistently learns the same path regardless of $\xi$. Although both agents easily solve the problem, the variance of the randomized agent's policy can be problematic in more complex environments in which identifying similarities between domains and ignoring irrelevant differences is important.

Next, we compare the measured difference between the policies learned by regularized agents on the two domains to the smallest of our theoretical bounds in equation \ref{proposition}, which in this simple environment can be directly calculated. For a given value of $\lambda$, we train a regularized agent on the reference domain. We then measure the difference in returns obtained by this agent on the reference and on the regularized domain, and this return determines the agent’s position along the $y$ axis. We then numerically calculate the Lipschitz constant from the agent’s action distribution over all states, and use this constant to calculate the bound in proposition 1. This bound determines the agent’s position along the $x$ axis. Our results for different random seeds and values of $\lambda$ are shown in figure \ref{fig:gridworld}. We observe that increasing $\lambda$ does lead to decreases in both the empirical difference in returns and in the theoretical bound.

\section{Experimental details}

The code used for our experiments is available at \url{https://github.com/IndustAI/visual-domain-randomization}.

\subsection{State preprocessing}

For our implementation of the visual cartpole environment, each image consists of $84\times84$ pixels with RGB channels. To include momentum information in our state description, we stack $k=3$ frames, so the shape of the state that is sent to the agent is $84\times84 \times 9$.

We note that because of this preprocessing, agents trained until convergence achieve average returns of about 175 instead of the maximum achievable score of 200. Since the raw pixels do not contain momentum information, we stack three frames as input to the network. When the environment is reset, we thus choose to perform two random actions before the agent is allowed to make a decision. For some initializations, this causes the agent to start in a situation it cannot recover from. Moreover, due to the low image resolution the agent may sometimes struggle to correctly identify momentum and thus may make mistakes.

In CarRacing, each state consists of $96\times96$ pixels with RGB channels. We introduce frame skipping as is often done for Atari games (\cite{mnih2015human}), with a skip parameter of 5. This restricts the length of an episode to 200 action choices. We then stack 2 frames to include momentum information into the state description. The shape of the state that is sent to the agent is thus $96\times96 \times 6$. We note that although this preprocessing makes training agents faster, it also causes trained agents to not attain the maximum achievable score on this environment.

\subsection{Visual Cartpole}

\subsubsection{Representations learned by the agents}

\begin{figure}[ht]
\centering
\includegraphics[scale=0.24]{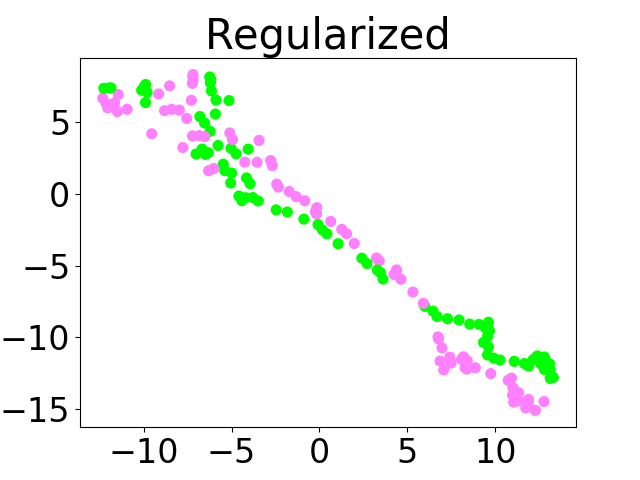}
\includegraphics[scale=0.24]{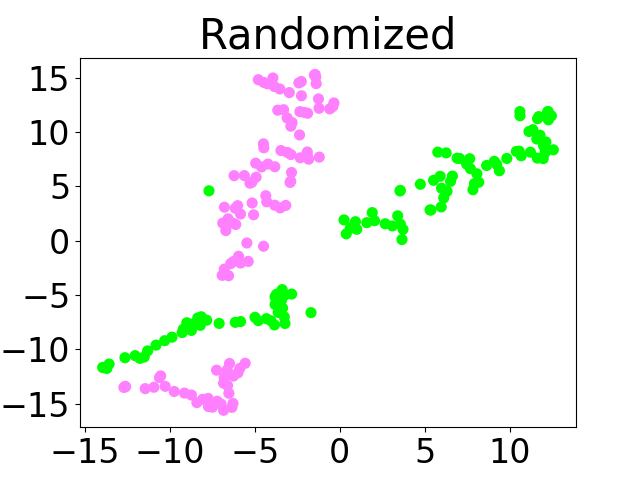}
\raisebox{1.1\height}{
\begin{tabular}{|c|c|}
  \hline
  Agent & Standard \\ 
  & Deviations \\
  \hline
  Normal & 10.1 \\
  Randomized & 6.2 \\
  Regularized (ours) & \textbf{3.7} \\
  \hline
\end{tabular}}
\caption{Left: Visualization of the representations learned by the agents for pink and green background colors and for the same set of states. We observe that the randomized agent learns different representations for the two domains. Right: Standard deviation of estimated value functions over randomized domains, averaged over 10 training seeds.}
\label{fig:tsne}
\end{figure}

To understand what causes the difference in behavior between the regularized and randomized agents on Cartpole, we study the representations learned by the agents by analyzing the activations of the final hidden layer. We consider the agents trained on $\{\phi\}_{big}$, and a sample of states obtained by performing a greedy rollout on a white background (which is included in $\{\phi\}_{big}$). For each of these states, we calculate the representation corresponding to that state for another background color in $\{\phi\}_{big}$.  We then visualize these representations using t-SNE plots, where each color corresponds to a domain. A representative example of such a plot is shown in figure \ref{fig:tsne}. We see that the regularized agent learns a similar representation for both backgrounds, whereas the randomized agent clearly separates them. This result indicates that the regularized agent learns to ignore the background color, whereas the randomized agent is likely to learn a different policy for a different background color. Further experiments comparing the representations of both agents can be found in the appendix.

To quantitatively study the effect of our regularization method on the representations learned by the agents, we compare the variations in the estimated value function for both agents over $\{\phi\}_{big}$. Figure \ref{fig:tsne} shows the standard deviation of the estimated value function over different background colors, averaged over 10 training seeds and a sample of states obtained by the same procedure as described above. We observe that our regularization technique successfully reduces the variance of the value function over the randomization domain.

\subsubsection{Extrapolation}

\begin{figure}[h]
\centering
\includegraphics[scale=0.3]{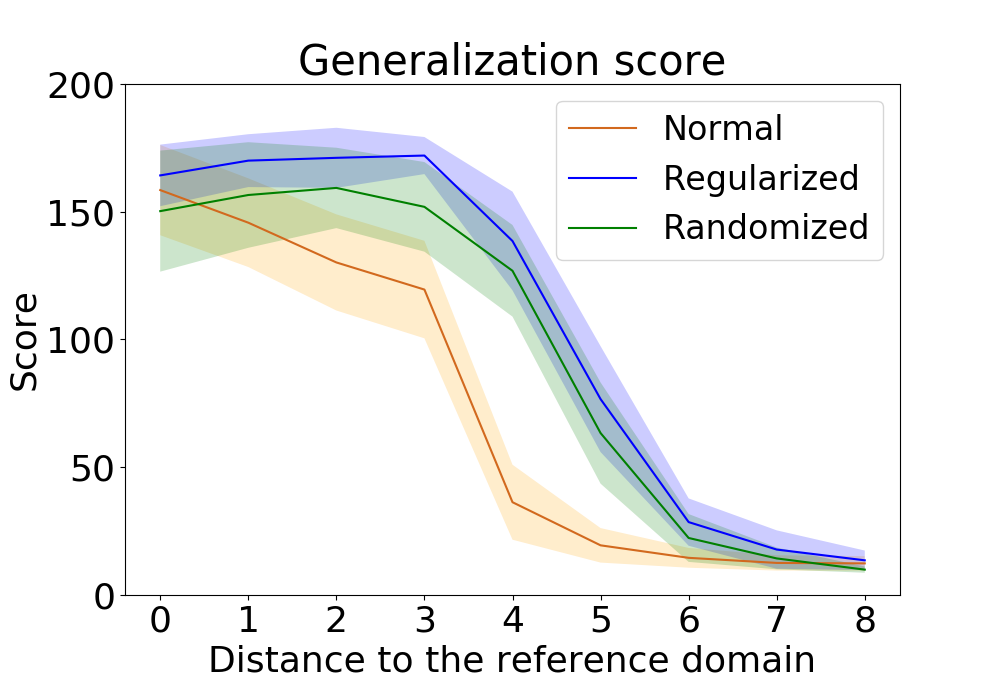}
\caption{Generalization scores, with 95\% confidence intervals obtained over 10 training seeds. The normal agent is trained on white $(1,1,1)$, corresponding to a \textit{distance to train}$ =0$. The rest of the domains correspond to $(x,x,x)$, for $x= 0.9, 0.8, \dots, 0.$}
\label{fig:generalization}
\end{figure}

Given that regularized agents are stronger in interpolation over their training domain, it is natural to wonder what the performance of these agents is in extrapolation to colors not within the range of colors sampled within training. For this purpose, we consider randomized and regularized agents trained on $\{\phi\}_{big}$, and test them on the set $\{(x,x,x), 0\leq x\leq 1\}$. None of these agents was ever exposed to $x\leq 0.5$ during training. 

Our results are shown in figure \ref{fig:generalization}. We find that although the regularized agent consistently outperforms the randomized agent in interpolation, both agents fail to extrapolate well outside the train domain. Since we only regularize with respect to the training space, there is indeed no guarantee that our regularization method can produce an agent that extrapolates well. Since the objective of domain randomization often is to achieve good transfer to an a priori unknown target domain, this result suggests that it is important that the target domain lie within the randomization space, and that the randomization space be made as large as possible during training.

\subsubsection{Further study of the representations learned by different agents}

\begin{figure}[h]
\centering
\includegraphics[scale=0.3]{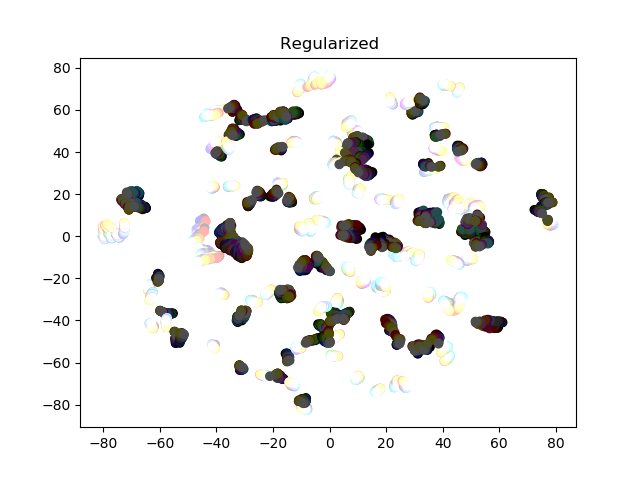}
\includegraphics[scale=0.3]{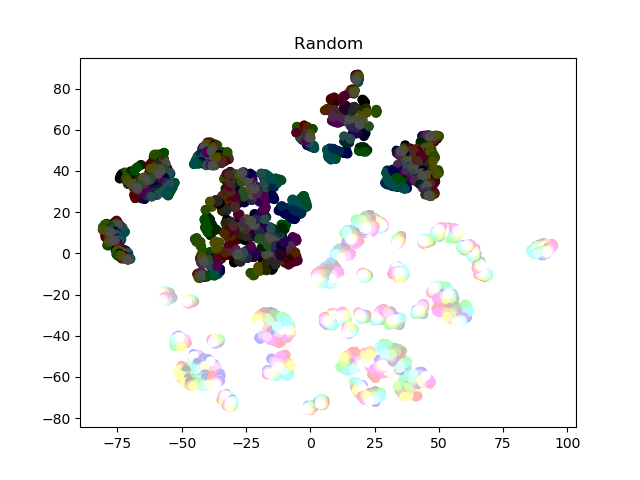}
\caption{t-SNE of the representations over $\{\phi\}_{split}$ of the Regularized (Left) and Randomized (Right) agents. Each color corresponds to a domain. The randomized agent learns very different representations for $[0,0.2]^3$ and $[0.8,1]^3$.}
\label{fig:tsne_split}
\end{figure}

We perform further experiments to demonstrate that the randomized agent learns different representations for different domains, whereas the regularized agent learns similar representations. We consider agents trained on $\{\phi\}_{split} = [0,0.2]^3 \cup [0.8,1]^3$, the union of \textit{darker}, and \textit{lighter} backgrounds. We then rollout each agent on a single episode of the domain with a white background and, for each state in this episode, calculate the representations learned by the agent for other background colors. We visualize these representations using the t-SNE plot shown in figure \ref{fig:tsne_split}. We observe that the randomized agent clearly separates the two training domains, whereas the regularized agent learns similar representations for both domains.

\begin{figure}[h]
\centering
\includegraphics[scale=0.3]{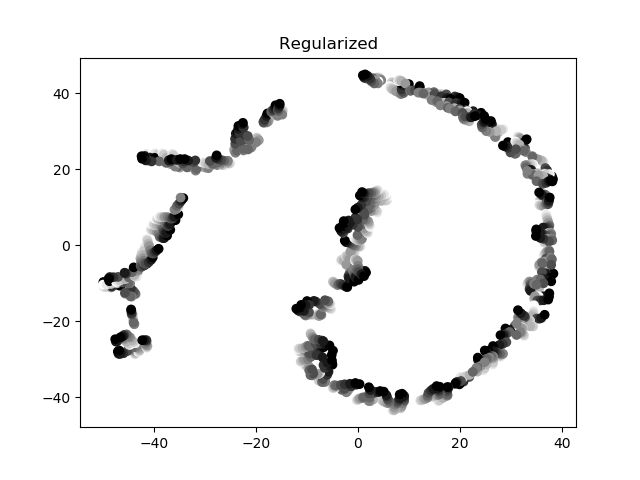}
\includegraphics[scale=0.3]{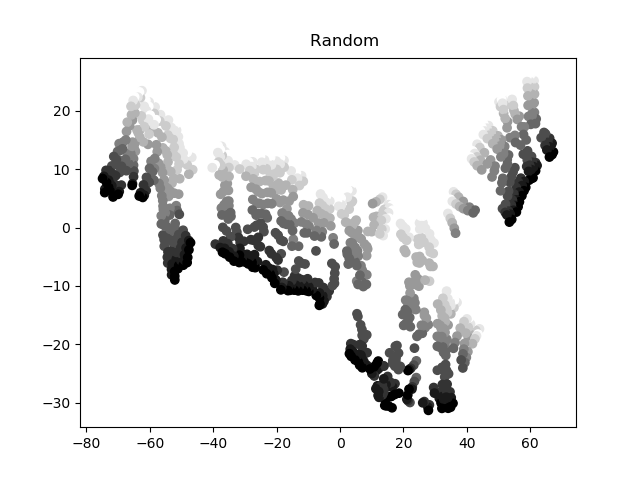}
\caption{t-SNE of the features of the Regularized (Left) and Randomized (Right) agents. Each color corresponds to a domain. }
\label{fig:tsne_xxx}
\end{figure}

We are interested in how robust our agents are to unseen randomizations $\phi \not\in \{\phi\}_{split}$. To visualize this, we rollout both agents in domains having different background colors : $\{(x,x,x), 0\leq x\leq 1\}$, i.e ranging from black to white, and collect their features over an episode. We then plot the t-SNEs of these features for both agents in figure \ref{fig:tsne_xxx}, where each color corresponds to a domain.

We observe once again that the regularized agent has much lower variance over unseen domains, whereas the randomized agent learns different features for different domains. This shows that the regularized agent is more robust to domain shifts than the randomized agent.

\section{Algorithms}

\begin{algorithm}[H]
   \caption{Deep Q-learning with our regularization method}
\begin{algorithmic}
   \STATE Initialize replay memory $\mathcal{D}$ to capacity $N$
   \STATE Initialize action-value function $Q$ with random weights $\theta$
   \STATE Initialize the randomization space $\{\phi\}$, and a reference MDP $M_{\phi^{ref}}$ to train on.
   \STATE Initialize a regularization parameter $\lambda$
   \STATE Define a feature extractor $f_\theta$
   \FOR {episode = $1, M$}
			\STATE Sample a randomizer function $\phi^{sampled}$ uniformly from $\{\phi\}$.
			\FOR {$t = 1, T$}
				\STATE With probability $\epsilon$ select a random action $a_t$
				\STATE otherwise select $a_t = \argmax_a Q(\phi^{ref}(s_t), a; \theta)$
				\STATE Execute action $a_t$ in $M_{\phi^{ref}}$, observe reward $r_t$, the reference state, and the corresponding randomized state with the chosen visual settings:  $\phi^{ref}(s_{t+1}), \phi^{sampled}(s_{t+1})$
				\STATE Store transition $(\phi^{ref}(s_{t}), \phi^{sampled}(s_{t}), a_t, r_t,\phi^{ref}(s_{t+1}), \phi^{sampled}(s_{t+1}))$ in $\mathcal{D}$
				\STATE Sample random minibatch of transitions $(\phi^{ref}_j, \phi^{sampled(old)}_j, a_j, r_j, \phi^{ref}_{j+1}, \phi^{sampled(old)}_{j+1})$ from $\mathcal{D}$, where $\phi^{sampled(old)}_j$ is the randomization that had been selected when the transition had been observed.
				\STATE Set $ y_j =
									   \begin{cases}
									     & r_j       \text{for terminal } \phi^{ref}_{j+1}.\\
									     &  r_j + \gamma \max_{a'} Q(\phi^{ref}_{j+1}, a', \theta) \\ &  \text{otherwise. }
									   \end{cases}
									 $
				\STATE Perform a gradient descent step on $(y_j - Q(\phi^{ref}_j, a_j; \theta))^2 +\lambda \|f_\theta(\phi^{ref}_j) - f_\theta(\phi^{sampled(old)}_j)\|_2^2$
				
			\ENDFOR
		\ENDFOR
\end{algorithmic}
\end{algorithm}
\newpage
\begin{algorithm}[H]
   \caption{Policy Gradient with a baseline using our regularization method}
\begin{algorithmic}
   \STATE Initialize policy network function $\pi_\theta$ with random weights $\theta$, baseline $b_\theta$
   \STATE Initialize the randomization space $\{\phi\}$, and a reference MDP $M_{\phi^{ref}}$ to train on.
   \STATE Initialize a regularization parameter $\lambda$
   \STATE Define a feature extractor $f_\theta$
   \FOR {episode = $1, M$}
			\STATE Collect a set of trajectories $(s_0,a_0,r_1, \dots, s_{T-1},$ \\ $ a_{T-1}, r_T)$ by executing $\pi_\theta$ on $M_{\phi^{ref}}$.
			\STATE For each trajectory, sample a randomizer function $\phi^{sampled}$ uniformly from $\{\phi\}$.
			\FOR {$t = 1, T$ in each trajectory }  
				\STATE Compute the return $R_t = \sum_{t'=t}^{T-1}\gamma^{t'-t}r_{t'}$
				\STATE Estimate the advantage $\hat{A_t} = R_t-b_\theta(\phi^{ref}(s_t))$

			\ENDFOR
			\STATE Perform a gradient descent step on
			\begin{align*}
			    &\sum_{t= 0}^T\big[-\hat{A_t}\log\pi_\theta(a_t|\phi^{ref}(s_t)) \\&+ \|R_t - b_\theta(\phi^{ref}(s_t))\|_2^2 \\&+\lambda \|f_\theta(\phi^{ref}(s_t)) - f_\theta(\phi^{sampled}(s_t))\|_2^2 \big]
		    \end{align*}
		\ENDFOR
\end{algorithmic}
\label{alg:policy}
\end{algorithm}

Note that algorithm \ref{alg:policy} can be straightforwardly adapted to several state of the art policy gradient algorithms such as PPO.

\end{document}